\RequirePackage{fix-cm}
\documentclass[twocolumn]{svjour3}          
\smartqed  
\usepackage{booktabs}
\usepackage{amsmath}
\usepackage{amssymb}
\usepackage{mathtools}
\usepackage{xspace}
\usepackage{hyperref}
\usepackage{graphicx}
\usepackage{caption}
\usepackage{subcaption}
\usepackage{multirow}
\usepackage{color, colortbl}
\usepackage{fancyvrb}
\usepackage{blindtext}
\usepackage{appendix}
\usepackage{natbib}

\usepackage{etoolbox}
\makeatletter
\patchcmd\@combinedblfloats{\box\@outputbox}{\unvbox\@outputbox}{}{%
   \errmessage{\noexpand\@combinedblfloats could not be patched}%
}%
 \makeatother

\definecolor{Gray}{gray}{0.9}
\definecolor{pink}{rgb}{1.0, 0.13, 0.32}

\newcommand{\eg}{\emph{e.g.,}\xspace}

\journalname{International Journal of Computer Vision}

\hypersetup{pdfstartview={FitH}}
\hypersetup{pdfborder={0 0 0}}
\hypersetup{pdfpagemode={UseNone}}
\hypersetup{colorlinks=true}
\hypersetup{citecolor=blue}
\hypersetup{linkcolor=blue}

\begin{document}
\begin{sloppypar}

\title{ Pixelated Semantic Colorization}


\author{Jiaojiao~Zhao \and Jungong~Han \and Ling~Shao          \and Cees~G.~M.~Snoek 
}


\institute{Jiaojiao Zhao \at
              Universiteit van Amsterdam, Amsterdam, the Netherlands\\
              \email{j.zhao3@uva.nl}           
           \and
           Jungong Han \at
              Lancaster University, Lancaster, UK\\
              \email{jungonghan77@gmail.com} 
           \and
           Ling Shao \at
              Inception Institute of Artificial Intelligence, Abu Dhabi, UAE\\
              \email{ling.shao@ieee.org} 
           \and
           Cees G. M. Snoek \at
              Universiteit van Amsterdam, Amsterdam, the Netherlands\\
              \email{cgmsnoek@uva.nl}
}

\date{Received: date / Accepted: date}

\maketitle

\begin{abstract}
While many image colorization algorithms have recently shown the capability of producing plausible color versions from gray-scale photographs, they still suffer from limited semantic understanding. To address this shortcoming, we propose to exploit pixelated object semantics to guide image colorization. The rationale is that human beings perceive and distinguish colors based on the semantic categories of objects. Starting from an autoregressive model, we generate image color distributions, from which diverse colored results are sampled. We propose two ways to incorporate object semantics into the colorization model: through a pixelated semantic embedding and a pixelated semantic generator. Specifically, the proposed network includes two branches. One branch learns what the object is, while the other branch learns the object colors. The network jointly optimizes a color embedding loss, a semantic segmentation loss and a color generation loss, in an end-to-end fashion. Experiments on PASCAL VOC2012 and COCO-stuff reveal that our network, when trained with semantic segmentation labels, produces more realistic and finer results compared to the colorization state-of-the-art.
\end{abstract}

\begin{figure*}[t!]
\centering
\includegraphics[scale=0.7]{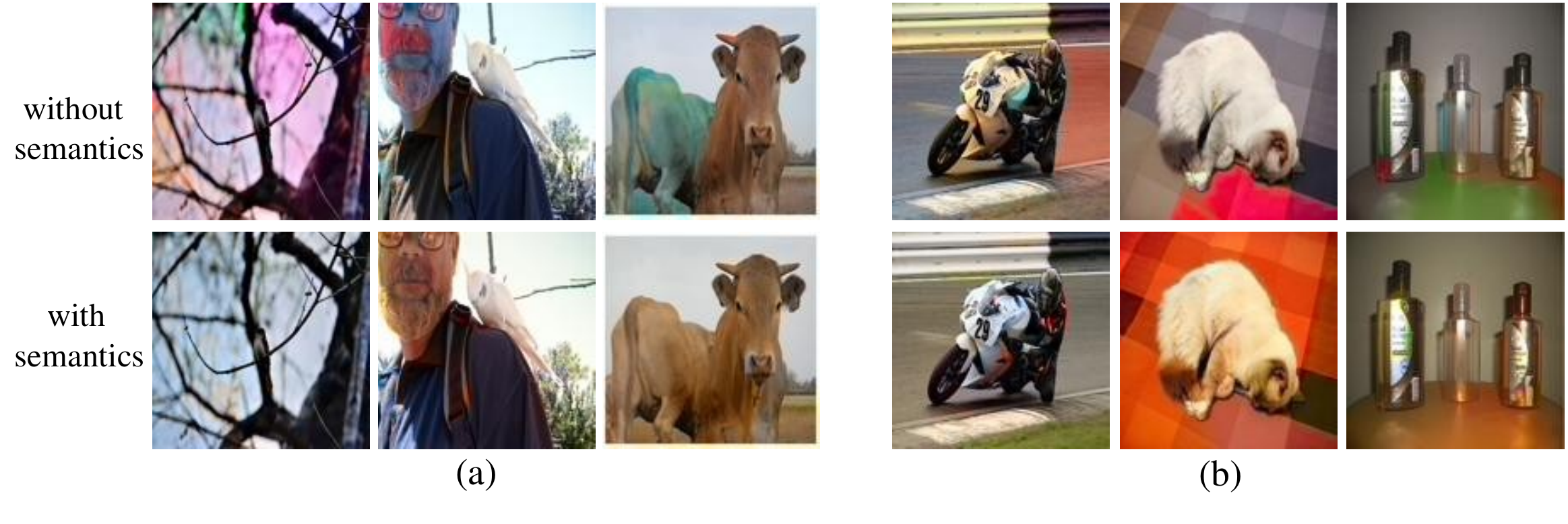}
\caption{\textbf{Colorization without and with semantics} generated using the network from this paper. (a) The method without semantics assigns unreasonable colors to objects, such as the colorful sky and the blue cow. The method with semantics generates realistic colors for the dog and the old man. (b) The method without semantics fails to capture long-range pixel interactions (\cite{ame2017bmvc}). With semantics, the model performs better.  
}
\label{Fig:samples}
\end{figure*}

\section{Introduction}
Color has been at the center stage of computer vision for decades, \eg (\cite{swain91,comaniciu1997robust,perez2002color,khan2009top,SandePAMI10,lou2015color,vondrick2018tracking}). Many vision challenges, including object detection and visual tracking, benefit from color (\cite{khan2009top,khan2012color,danelljan2014adaptive,vondrick2018tracking}). Consequently, color constancy (\cite{GijsenijIJCV10}) and color correction (\cite{sanchez2000improving}) methods may further enhance visual recognition . Likewise, color is commonly added to gray-scale images to increase their visual appeal and perceptually enhance their visual content, \eg (\cite{welsh2002transferring,iizuka2016let,zhang2016colorful,ame2017bmvc,deshpande2017learning}). This paper is about image colorization.

Human beings excel in assigning colors to gray-scale images since they can easily recognize the objects and have gained knowledge about their colors.  No one doubts the sea is typically blue and a dog is never naturally green. Although many objects have diverse colors, which makes their prediction quite subjective, humans can get around this by simply applying a bit of creativity. However, it remains a significant challenge for machines to acquire both the world knowledge and ``imagination'' that humans possess.   

Previous works in image colorization require reference images (\cite{gupta2012image,liu2008intrinsic,charpiat2008automatic}) or color scribbles (\cite{levin2004colorization}) to guide the colorization.  Recently, several automatic approaches (\cite{iizuka2016let,larsson2016learning,zhang2016colorful,ame2017bmvc,guadarrama2017pixcolor}) have been proposed based on deep convolutional neural networks. Despite the improved colorization, there are still common pitfalls that make the colorized images appear less realistic. We show some examples in Figure~\ref{Fig:samples}. The cases in (a) without semantics suffer from  incorrect semantic understanding. For instance, the cow is assigned a blue color. The cases in (b) without semantics suffer from color pollution. Our objective is to effectively address both problems to generate better colorized images with high quality.

Both traditional (\cite{chia2011semantic,ironi2005colorization}) and recent colorization solutions (\cite{larsson2016learning,iizuka2016let,he2016deep,zhang2016colorful,zhang2017real}) have highlighted the importance of semantics. However, they only explore image-level class semantics for colorization. As stated by \cite{dai2016r}, image-level classification favors translation invariance.  Obviously, colorization requires representations that are, to a certain extent, translation-variant. From this perspective, semantic segmentation (\cite{long2015fully,chen2018deeplab,noh2015learning}), which also requires translation-variant representations, provides more reasonable semantic guidance for colorization. It predicts a class label for each pixel. Similarly, according to (\cite{zhang2016colorful,larsson2016learning}), colorization assigns each pixel a color distribution. Both challenges can be viewed as an image-to-image prediction problem and formulated as a pixel-wise prediction task. We show several colorized examples after using pixelated semantic-guidance in Figure~\ref{Fig:samples} (a) and (b). Clearly, pixelated semantics helps to reduce the color inconsistency by a better semantic understanding.

In this paper, we study the relationship between colorization and semantic segmentation. Our proposed network is able to be harmoniously trained for semantic segmentation and colorization. By using such multi-task learning, we explore how pixelated semantics affects colorization. Differing from the preliminary conference version of this work (\cite{zhao2018pixel}), we view colorization here as a sequential pixel-wise color distribution generation task, rather than a pixel-wise classification task. We design two ways to exploit pixelated semantics for colorization, one by guiding a
color embedding function and the other by guiding a color generator. 
Using these strategies, our methods produce diverse vibrant images on two datasets, Pascal VOC2012 (\cite{everingham2015pascal}) and COCO-stuff (\cite{caesar2018cvpr}). We further study how colorization can help semantic segmentation and demonstrate that the two tasks benefit each other. We also propose a new quantitative evaluation method using semantic segmentation accuracy. 

The rest of the paper is organized as follows. In Section 2, we introduce related work. Following, in Section 3, we describe the details of our colorization network using pixelated semantic guidance. Experiments and results are presented in Section 4. We conclude our work in section 5.

\section{Related Work}

\subsection{Colorization by Reference}
 Colorization using references was first proposed by \cite{welsh2002transferring}, who transferred the colors by matching the statistic within the pixel's neighborhood. Rather than relying on independent pixels, \cite{ironi2005colorization} transferred colors from a segmented example image based on their observation that pixels with the same luminance value and similar neighborhood statics may appear in different regions of the reference image, which may have different semantics and colors. \cite{tai2005local} and \cite{chia2011semantic} also performed local color transfer by segmentation. \cite{bugeau2014variational} and \cite{gupta2012image} proposed to transfer colors at pixel level and super-pixel level. Generally, finding a good reference with similar semantics is key for this type of methods. Previously, \cite{liu2008intrinsic} and \cite{chia2011semantic} relied on image retrieval methods to choose good references. Recently, deep learning has supplied more automatic methods in (\cite{cheng2015deep,he2018deep}). In our approach, we use a deep network to learn the semantics from data, rather than relying on a reference with similar semantics.

\subsection{Colorization by Scribble}
Another interactive way to colorize a gray-scale image is by placing scribbles. This was first proposed by \cite{levin2004colorization}. The authors assumed that pixels nearby in space-time, which have similar gray levels, should have similar colors as well. Hence, they solved an optimization problem to propagate sparse scribble colors. To reduce color bleeding over object boundaries, \cite{huang2005adaptive} adopted an adaptive edge detection to extract reliable edge information. \cite{qu2006manga} colorized manga images by propagating scribble colors within the pattern-continuous regions. \cite{yatziv2006fast} developed a fast method to propagate scribble colors based on color blending. \cite{luan2007natural}, further extended \cite{levin2004colorization} by grouping not only neighboring pixels with similar intensity but also remote pixels with similar texture. Several more current works (\cite{zhang2017real,sangkloy2017scribbler}) used deep neural networks with scribbles trained on a large dataset and achieved impressive colorization results. In all these methods, which use hints like strokes or points, provide an important means for segmenting an image into different color regions. We prefer to learn the segmentation rather than manually labelling it.   

\begin{figure*}[t!]
\centering
\includegraphics[scale=0.6]{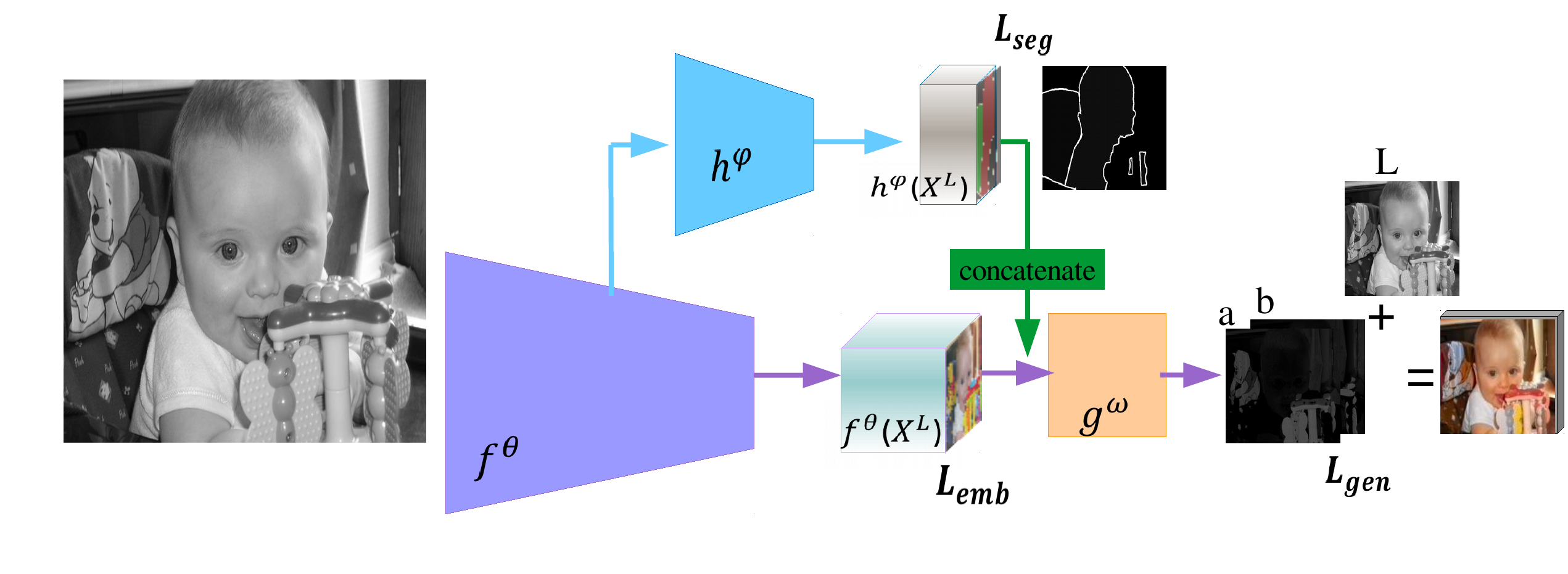}
\caption{\textbf{Pixelated semantic colorization}. The three colored flows (arrows) represent three variations of our proposal. The purple flow illustrates the basic pixelated colorization backbone (Section 3.1). The purple flow combined with the blue flow obtains a better color embedding with more semantics (Section 3.2.1). The purple flow, blue flow and green flow together define our final model, a pixelated colorization model conditioned on gray-scale image and semantic labels (Section 3.2.2). Here, $f^\theta$ is a color embedding function, $h^\varphi$ is a semantic segmentation head and $g^\omega$ is the autoregressive generation model. There are three loss functions $L_{seg}$, $L_{emb}$ and $L_{gen}$ (Section 3.3).}  
\label{Fig:method}
\end{figure*}

\subsection{Colorization by Deep Learning}
The earliest work applying a deep neural network was proposed by \cite{cheng2015deep}. They first grouped images from a reference database into different clusters and then learned deep neural networks for each cluster. Later, \cite{iizuka2016let}
pre-trained a network on ImageNet for a classification task, which provided global semantic supervision. The authors leveraged a large-scale scene classification database to train a model, exploiting the class-labels of the dataset to learn the global priors. Both of these works treated colorization as a regression problem. In order to generate more saturated results, \cite{larsson2016learning} and \cite{zhang2016colorful} modeled colorization as a classification problem. \cite{zhang2016colorful} applied cross-channel encoding as
self-supervised feature learning with semantic interpretability. \cite{larsson2016learning} 
claimed that interpreting the semantic composition of the scene and localizing objects were key to colorizing arbitrary images. Nevertheless, these works only explored image-level classification semantics. Our method takes the semantics one step further and utilizes finer pixelated semantics from segmentation.

Further, generative models have more recently been applied to produce diverse colorization results. Currently, several works (\cite{cao2017unsupervised,isola2017image,frans2017outline}) have applied a generative adversarial network (GAN) (\cite{radford2015unsupervised}). They were able to produce sharp results but were not as good as the approach proposed by \cite{zhang2016colorful}. Variational autoencoders (VAE) (\cite{kingma2013auto}) have also been used to learn a color embedding (\cite{deshpande2017learning}). This method produced results with large-scale spatial co-ordination but tonelessness. \cite{ame2017bmvc} and \cite{guadarrama2017pixcolor} applied PixelCNN (\cite{van2016conditional,salimans2017pixelcnn++}) to generate better results. We use PixelCNN as the backbone in this paper.

\section{Methodology}
In this section, we will detail how pixelated semantics improves colorization. We will first introduce our basic colorization backbone. Then, we will present two ways to exploit object semantics for colorization. Our network structure is summarized in Figure~\ref{Fig:method}.    

\subsection{Pixelated Colorization}
To arrive at image colorization with pixelated semantics, we start from an autoregressive model. It colorizes each pixel conditioned on the input gray image and previously colored pixels. Specifically, a conditional PixelCNN (\cite{van2016conditional}) is utilized to generate per-pixel color distributions, from which we sample diverse colorization results.

We rely on the CIE \textbf{Lab} color space to perform the colorization, since it was designed to be perceptually uniform with respect to human color vision and only two channels \textbf{a} and \textbf{b} need to be learned. An image with a height $H$ and a width $W$ is defined as $X \in R^{H\times W \times3}$. $X$ contains $n$ $(=H\times W)$ pixels. In raster scan order: row by row and pixel by pixel within every row, the value of the $i$-th pixel is denoted as $X_i$. The input gray-scale image, represented by light channel \textbf{L}, is defined as $X^L\in R^{H\times W \times1}$. The objective of colorization is to predict the \textbf{a} and \textbf{b} channels $\hat{Y}\in R^{H\times W \times2}$.  Different from the \textbf{RGB} color space, \textbf{Lab} has the range $[0;100]\times[-127;128]\times[-127;128]$.

To reduce computation and memory requirements, we prefer to produce color images with low resolution. This is reasonable since the human visual system resolves color less precisely than intensity (\cite{van1969spatiotemporal}). As stated in (\cite{ame2017bmvc}), image compression schemes, such as JPEG, or previously proposed techniques for automatic colorization also apply chromatic subsampling. 

By adopting PixelCNN for image colorization, a joint distribution with condition is modelled as~\cite{van2016conditional}:
\begin{equation}
     p(\hat{Y}|X^L)=\mathop{\prod}\limits_{i=1}^{n}p(\hat{Y}_i|\hat{Y}_1,...,\hat{Y}_{i-1},X^L).   
\end{equation} 
All the elementary per-pixel conditional distributions are modelled using a shared convolutional neural network. As all variables in the factors are observed, training can be executed in parallel. 

Furthermore, $X^L$ can be replaced by a good embedding learned from a neural network. Taking $g^\omega$ as the generator function and $f^\theta$ as the embedding function, each distribution in Equation (1) can be rewritten as:
\begin{equation}
     p(\hat{Y}_i|\hat{Y}_1,...,\hat{Y}_{i-1},X^L) = g^\omega_i(\hat{Y}_1,...,\hat{Y}_{i-1},f^\theta(X^L)).   
\end{equation}

As the purple flow in Figure~\ref{Fig:method} shows, there are two components included in our model. A deep convolutional neural network ($f_\theta$) produces a good embedding of the input gray-scale image. Then an autoregressive model uses the embedding to generate a color distribution for each pixel. The final colorized results are sampled from the distributions using a pixel-level sequential procedure. We first sample $\hat{Y}_1$ from $p(\hat{Y}_1|X^L)$, then sample $\hat{Y}_i$ from $p(\hat{Y}_i|\hat{Y}_1,...,\hat{Y}_{i-1},X^L)$ for all $i$ in ${\{2,...n\}}$.

\subsection{Pixelated Semantic Colorization}
Intuitively, semantics is the key to colorizing objects and scenes. We will discuss how to embed pixelated semantics in our colorization model for generating diverse colored images. 

\subsubsection{Pixelated Semantic Embedding}

Considering the conditional pixelCNN model introduced above, a good embedding of gray-scale image $f^\theta(X^L)$ greatly helps to generate the precise color distribution of each pixel. We first incorporate semantic segmentation to improve the color embedding. We use $X^S$ to denote the corresponding segmentation map. Then, we learn an embedding of the gray-scale image conditioned on $X^S$. We replace $f^\theta(X^L)$ with $f^\theta(X^L|X^S)$. Thus, the new model learns the distribution in Equation (2) as:
\begin{equation}
\begin{aligned}
     p(\hat{Y}_i|\hat{Y}_1,...,\hat{Y}_{i-1},X^L,X^S) \\
     = g^\omega_i(\hat{Y}_1,...,\hat{Y}_{i-1},f^\theta(X^L|X^S)). 
\end{aligned}
\end{equation}
Here, the semantics only directly affects the color embedding generated from the gray-scale image, but not the autoregressive model.  

Incorporating semantic segmentation can be straightforward, i.e., using segmentation masks to guide the colorization learning procedure. Such a way enables the training phase to directly obtain guidance from the segmentation masks, which clearly and correctly contain semantic information. However, it is not suitable for the test phase as segmentation masks are needed. Naturally, we can rely on an off-the-shelf segmentation model to gain segmentation masks for all the test images, but it is not elegant. Instead, we believe it is best to simultaneously learn the semantic segmentation and the colorization, making the two tasks benefit each other, as we originally proposed in (\cite{zhao2018pixel}).

Modern semantic segmentation can easily share low-level features with the color embedding function. We simply need to plant an additional segmentation branch $h^\varphi$ following a few bottom layers, like the blue flow shown in Figure~\ref{Fig:method}. Specifically, we adopt the semantic segmentation strategies from \cite{chen2018deeplab}. At the top layer, we apply atrous spatial pyramid pooling, which expoits multiple scale features by employing multiple parallel filters with different rates. The final prediction ($h^\varphi(X^L)$) is the fusion of the features from the different scales, which helps to improve segmentation. The two tasks have different top layers for learning the high-level features. In this way, semantics is injected into the color embedding function. By doing so, a better color embedding with more semantic awareness is learned as input to the generator. This is illustrated in Figure~\ref{Fig:method}, by combining the purple flow and the blue flow.

\subsubsection{Pixelated Semantic Generator}
A good color embedding with semantics aids the generator to produce more correct color distributions. Furthermore, the generator is likely to be improved with semantic labels further. Here, we propose to learn a distribution conditioned on previously colorized pixels, a color embedding of gray-scale images with semantics ($f^\theta(X^L|X^S)$), and pixel-level semantic labels. We rewrite Equation (3) as:
\begin{equation}
\begin{aligned}
     p(\hat{Y}_i|\hat{Y}_1,...,\hat{Y}_{i-1},X^L,X^S) \\
     = g^\omega_i(\hat{Y}_1,...,\hat{Y}_{i-1},f^\theta(X^L|X^S), h^\varphi(X^L)). 
\end{aligned}
\end{equation}

Intuitively, this method is capable of using semantics
to produce more correct colors of objects and more continuous colors within one object. It is designed to address the two issues mentioned in Figure~\ref{Fig:samples}. The whole idea is illustrated in Figure~\ref{Fig:method} by combining the purple flow with the blue and green flows.

We consider two different ways to use pixelated semantic information to guide the generator. The first way is to simply concatenate the color embedding $f^\theta(X^L)$ and the segmentation prediction $h^\varphi(X^L)$ along the channels and then input the fusion to the generator. The second way is to apply a feature transformation introduced by~\cite{perez2017film} and ~\cite{wang2018recovering}. Specifically, we use convolutional layers to learn a pair of transformation parameters from the segmentation predictions. Then, a transformation is applied to the color embedding using these learned parameters. We find the first way works better. Results will be shown in the Experiment section.

\subsection{Networks}
In this section, we provide the details of the network structure and the optimization procedure. 

{\textbf{Network structure}}. Following the scheme in Figure~\ref{Fig:method}, three components are included: the color embedding function $f^\theta$, the semantic segmentation head $h^\varphi$ and the autoregressive model $g^\omega$. Correspondingly, three loss functions are jointly learned, which will be introduced later. The three flows represent the three different methods introduced above. The purple flow illustrates the basic pixelated colorization. The purple flow combined with the blue flow results in the pixelated semantic embedding. The purple flow combined with the blue and green flows, results in the pixelated semantic generator.

\begin{table}[t]
	\renewcommand\arraystretch{1.1}
	\centering
	\resizebox{\columnwidth}{!}{
		\begin{tabular}{c|c|c|c}
		\toprule
        \multicolumn{4}{c}{\textbf{Color embedding $f^\theta(X^L)$}} \\
         \cmidrule(lr){1-4}
			 Module & Resolution & Channels & Dilation \\
			\midrule
			\rowcolor{Gray}
          convolution $3\times3/1$  & 128 & 64 & -\\
          \rowcolor{Gray}
          Residual block $\times2$  & 128 & 64 & - \\
          	\rowcolor{Gray}
          convolution $3\times3/2$  & 64 & 128 & -\\
          \rowcolor{Gray}
          Residual block $\times2$  & 64 & 128 & - \\
          \rowcolor{Gray}
          convolution $3\times3/2$  & 32 & 256 & -\\
          \rowcolor{Gray}
          Residual block $\times2$  & 32 & 256 & - \\
          \rowcolor{Gray}
          convolution $3\times3/1$  & 32 & 512 & -\\
          \rowcolor{Gray}
          Residual block $\times3$  & 32 & 512 & 2 \\
          \hline

          convolution $3\times3/1$  & 32 & 512 & -\\

          Residual block $\times3$  & 32 & 512 & 4 \\

          convolution $3\times3/1$  & 32 & 160 & -\\ 
	\bottomrule
		\end{tabular}
		}
	\caption{\textbf{Color embedding branch structure}. Feature spatial resolution, number of channels and dilation rate are listed for each module. The gray rows indicate the bottom layers are shared with the semantic segmentation branch (detailed in Table~\ref{tab: Tab2}).}
	\label{tab: Tab1}
\end{table}

Inspired by the success of the residual block (\cite{he2016deep,chen2018deeplab}) and following \cite{ame2017bmvc}, we apply gated residual blocks (\cite{van2016conditional,salimans2017pixelcnn++}), each of which has two convolutions with $3\times3$ kernels, a skip connection and a gating mechanism. We apply atrous (dilated) convolutions to several layers to increase the network's field-of-view without reducing its spatial resolution. Table~\ref{tab: Tab1} and~\ref{tab: Tab2} list the details of the color embedding branch and the semantic segmentation branch, respectively. The gray rows are shared by the two branches.

{\textbf{Loss functions}}. During the training phase, we train the colorization and segmentation simultaneously. We try to minimize the negative log-likelihood of the probabilities:
\begin{equation}
\mathop{\arg\min}_{\theta,\varphi, \omega} \ \sum -\log p(\hat{Y}|f^\theta(X^L),h^\varphi(X^L)).    
\end{equation}
Specifically, we have three loss functions $L_{emb}$, $L_{seg}$ and $L_{gen}$ to train the color embedding, the semantic segmentation and the generator, respectively. The final loss function $L_{sum}$ is the weighted sum of these loss functions: 

\begin{equation}
L_{sum} = \lambda_1*L_{emb} + \lambda_2*L_{seg} + \lambda_3*L_{gen}.    
\end{equation}

Following~\cite{salimans2017pixelcnn++}, we use discretized mixture logistic distributions to approximate the distribution in Equation (3) and Equation (4). A mixture of 10 logistic distributions is applied. Thus, both $L^{emb}$ and $L^{gen}$ are discretized mixture logistic losses. 

As for semantic segmentation, generally it should be performed in the RGB image domain as colors are important for semantic understanding. However, the input of our network is a gray-scale image which is more difficult to segment. Fortunately, the network incorporating colorization learning supplies color information which in turn strengthens the semantic segmentation for gray-scale images. The mutual benefit among the three learning parts is the core of our network. It is also important to realize that semantic segmentation, as a supplementary means for colorization, is not required to be very precise. We use the cross entropy loss with the standard softmax function for semantic segmentation (\cite{chen2018deeplab}). 

\begin{table}[t]
	\renewcommand\arraystretch{1.1}
	\centering
	\resizebox{\columnwidth}{!}{
		\begin{tabular}{c|c|c|c}
		\toprule
        \multicolumn{4}{c}{\textbf{Semantic segmentation $h^\varphi(X^L)$}} \\
         \cmidrule(lr){1-4}
			 Module & Resolution & Channels & Dilation \\
			\midrule
			\rowcolor{Gray}
          convolution $3\times3/1$  & 128 & 64 & -\\
          \rowcolor{Gray}
          Residual block $\times2$  & 128 & 64 & - \\
          	\rowcolor{Gray}
          convolution $3\times3/2$  & 64 & 128 & -\\
          \rowcolor{Gray}
          Residual block $\times2$  & 64 & 128 & - \\
          \rowcolor{Gray}
          convolution $3\times3/2$  & 32 & 256 & -\\
          \rowcolor{Gray}
          Residual block $\times2$  & 32 & 256 & - \\
          \rowcolor{Gray}
          convolution $3\times3/1$  & 32 & 512 & -\\
          \rowcolor{Gray}
          Residual block $\times3$  & 32 & 512 & 2 \\

          \hline
          convolution $3\times3/1$  & 32 & 512 & -\\
          Residual block $\times3$  & 32 & 512 & 2 \\

          convolution $3\times3/1$  & 32 & \#class & 6\\ 
          convolution $3\times3/1$  & 32 & \#class & 12\\
          convolution $3\times3/1$  & 32 & \#class & 18\\
          add   & 32 & \#class & -\\
	\bottomrule
		\end{tabular}
		}
	\caption{\textbf{Semantic segmentation branch structure}. Feature spatial resolution, number of channels and dilation rate are listed for each module. \#class means the number of semantic categories. The gray rows indicate the bottom layers are shared with the color embedding branch (detailed in Table~\ref{tab: Tab1}).}
	\label{tab: Tab2}
\end{table} 


\section{Experiments}

\begin{figure}
\centering
\includegraphics[scale=0.8]{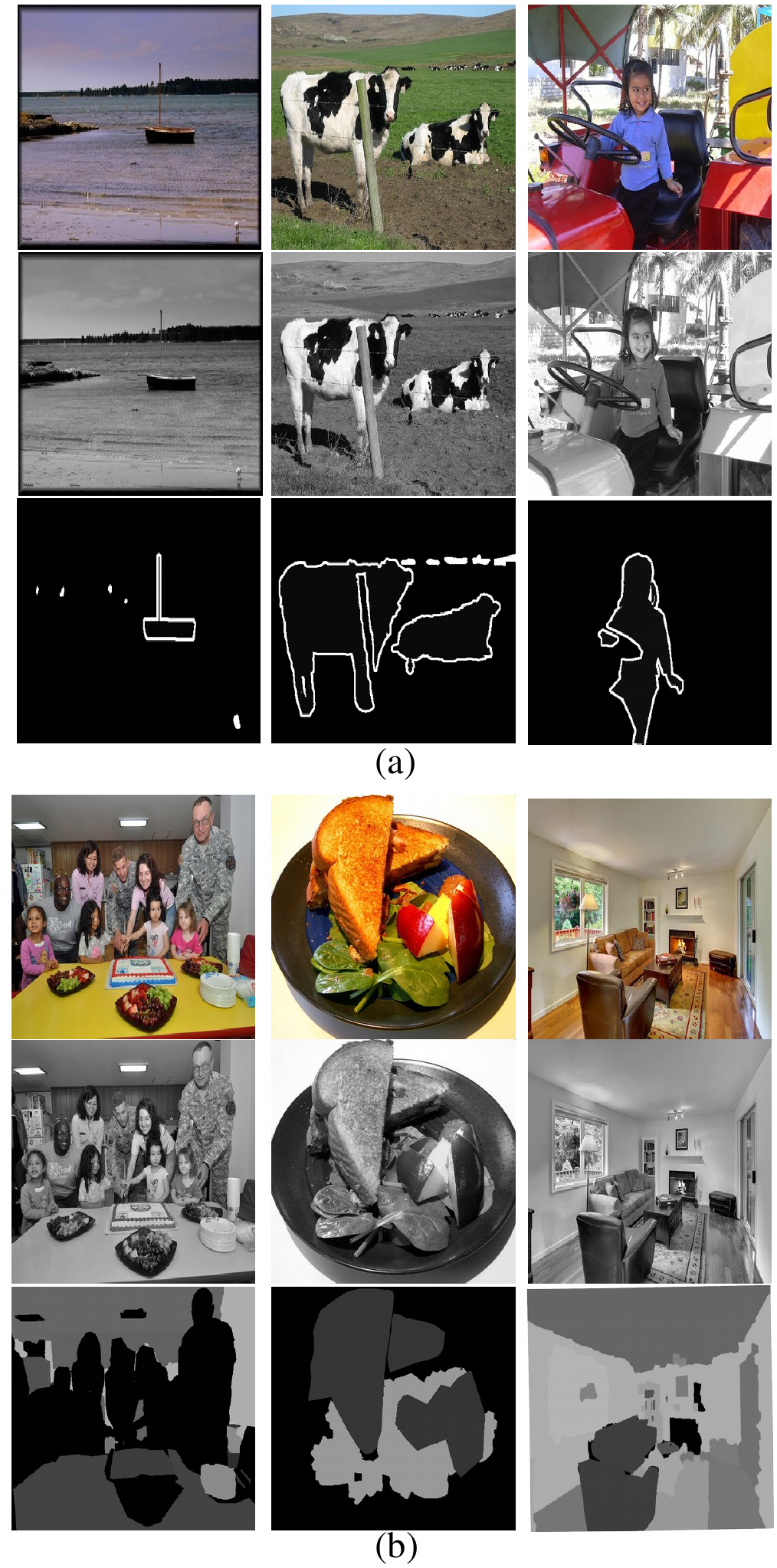}
\caption{\textbf{Color images, gray-scale images and segmentation maps} from (a) Pascal VOC and (b) COCO-stuff. COCO-stuff has more semantic categories than Pascal VOC.}  

\label{Fig:example}
\end{figure}

\begin{figure*}[t!]
\centering
\includegraphics[scale=0.85]{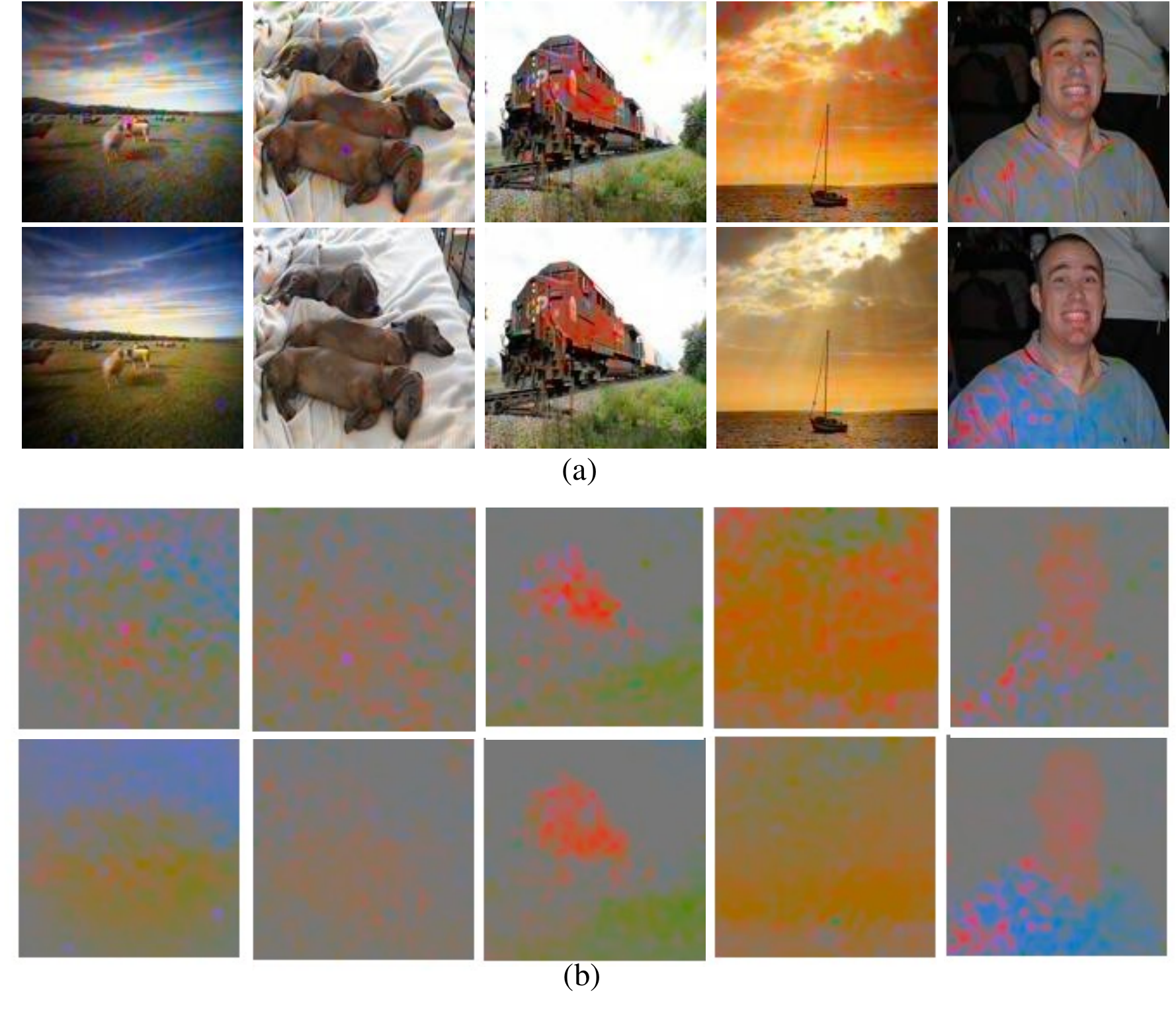}
\caption{{\textbf{Colorizations from the embedding functions $f^\theta$} using the purple flow and the purple-blue flow}. (a) Colorization without semantic-guidance (first row) and with semantic-guidance (second row). With semantics, better colorizations are produced. (b) Visualization of the predicted \textbf{a}
and \textbf{b} color channels of the colorizations. The top row shows the results without semantic-guidance and the bottom row with semantic-guidance. With semantics, the predicted colors have less noise and look more consistent.}   
\label{Fig:embed}
\end{figure*}

\begin{figure*}[t!]
\centering
\includegraphics[scale=0.65]{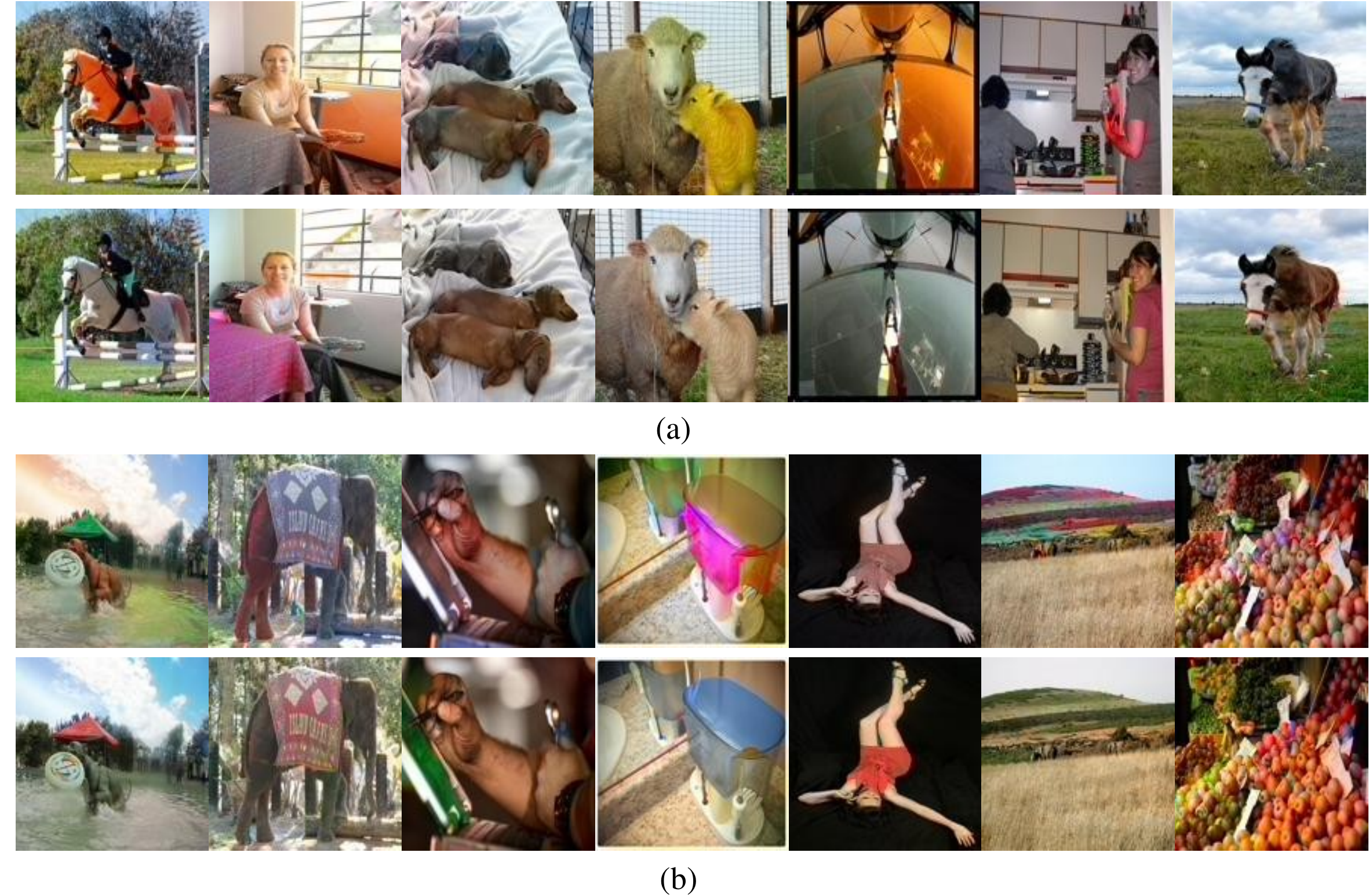}
\caption{\textbf{Colorization from the generators $g^\omega$}, when relying on the purple flow and the purple-blue flow. Examples from (a) Pascal VOC and (b) COCO-stuff are shown. For both datasets, the top row shows results from the model without semantic-guidance and the bottom row shows the ones with semantic-guidance. The results with semantic-guidance have more reasonable colors and better object consistency.}   
\label{Fig:result1}
\end{figure*}

\subsection{Experimental Settings}

{\textbf{Datesets}}. We report our experiments on Pascal VOC2012  (\cite{everingham2015pascal}) and COCO-stuff (\cite{caesar2018cvpr}). The former is a common semantic segmentation dataset with 20 object classes and one background class. Our experiments are performed on the 10582 images for training and the 1449 images in the validation set for testing. COCO-stuff is a subset of the COCO dataset (\cite{lin2014microsoft}) generated for scene parsing, containing 182 object classes and one background class on 9000 training images and 1000 test images. We use the gray-scale converted images as input and rescale each image to $128\times128$. Figure~\ref{Fig:example} shows some examples with natural scenes, objects and artificial objects from the datasets. 

{\textbf{Implementation}}. Commonly available pixel-level annotations intended for semantic segmentation are sufficient for our colorization method. We do not need new pixel-level annotations for colorization. We train our network with joint color embedding loss, semantic segmentation loss and generating loss with the weights $\lambda_1:\lambda_2:\lambda_3=1:100:1$, so that the three losses are similar in magnitude. Our multi-task learning for simultaneously optimizing colorization and semantic segmentation effectively avoids overfitting. The Adam optimizer (\cite{kingma2014adam}) is adopted. We set an initial learning rate equal to 0.001, momentum to 0.95 and second momentum to 0.9995. We apply Polyak parameter averaging (\cite{polyak1992acceleration}).

\subsection{Effect of segmentation on the embedding function $f^\theta$}
 We first study how semantic segmentation helps to improve the color embedding function $f^\theta$. Following the method introduced in Section 3.2.1, we jointly train the purple and blue flows shown in Figure~\ref{Fig:method}. In this case, the semantic segmentation branch only influences the color embedding function. To illustrate the effect of pixelated semantics, we compare the color embeddings generated from the embedding function $f^\theta$ in Figure~\ref{Fig:embed}. Obviously, as can be seen, semantic-guidance enables better color embeddings. For example, the sky in the first picture looks more consistent, and the sheep are assigned reasonable colors. However, the results without semantic-guidance appear less consistent. For instance, there is color pollution on the dogs and the sheet in the second picture.

Further, in order to more clearly show the predicted color channels of the color embeddings, we remove the light channel {\textbf{L}} and only visualize the chrominances {\textbf{a}} and {\textbf{b}}  in Figure~\ref{Fig:embed} (b). Interestingly, without semantic-guidance, the predicted colors are more noisy, as shown in the top row. However, with semantic-guidance, the colors are more consistent and echo the objects well. From these results, one clearly sees that colorization profits from semantic information. These comparisons support our idea and illustrate that pixelated semantics is able to enhance semantic understanding, leading to more consistent colorization.

\begin{figure*}[t!]
\centering
\includegraphics[scale=0.65]{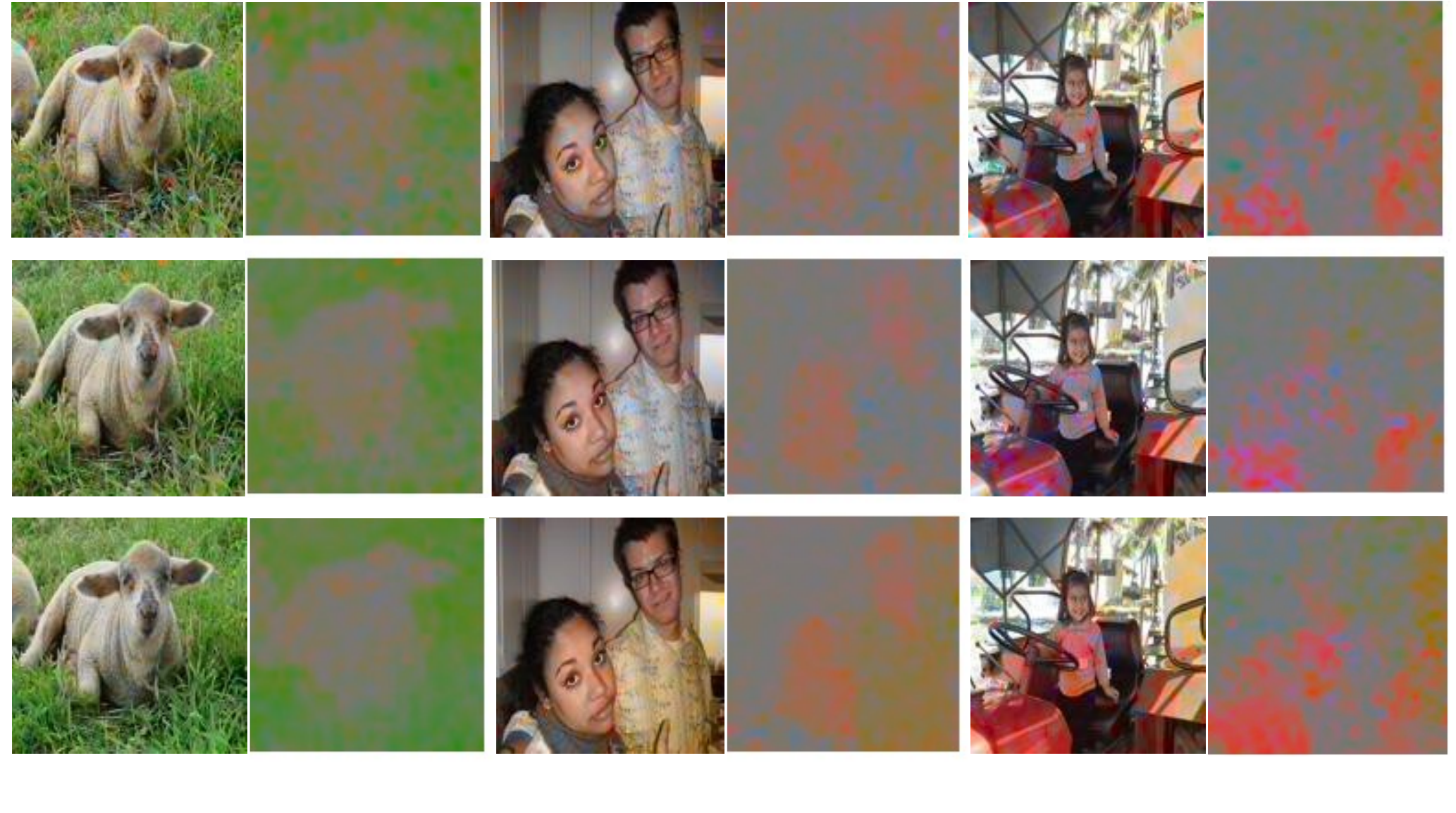}
\caption{\textbf{Colorizations generated by the embedding functions $f^\theta$, using three variants of our network}. The top row shows the results of the purple flow. The second row shows the results of the purple-blue flow. The bottom row shows the results of the purple-blue-green flow. Each colorization is followed by the corresponding predicted chrominances. The purple-blue-green flow produces the best colorization.}   
\label{Fig:embed2}
\end{figure*}

 In theory, we should obtain better colorization when a better color embedding is input into the generator. In Figure~\ref{Fig:result1}, we show some final colorizations produced by the generator $g^\omega$. Our method using pixelated semantics works well on the two datasets. The results look more realistic. For instance, the fifth example in the Pascal VOC dataset is a very challenging case. The proposed method generates consistent and reasonable color for the earth even with an occluded object. For the last example in Pascal VOC, it is surprising that the horse bit is assigned a red color although it is very tiny. The proposed method processes details well. We also show various examples from COCO-stuff, including animals, humans, fruits, and natural scenes. The model trained with semantics performs better. Humans are given normal skin color in the third and fifth examples. The fruits have uniform colors and look fresh.

\begin{figure*}[t!]
\centering
\includegraphics[scale=0.78]{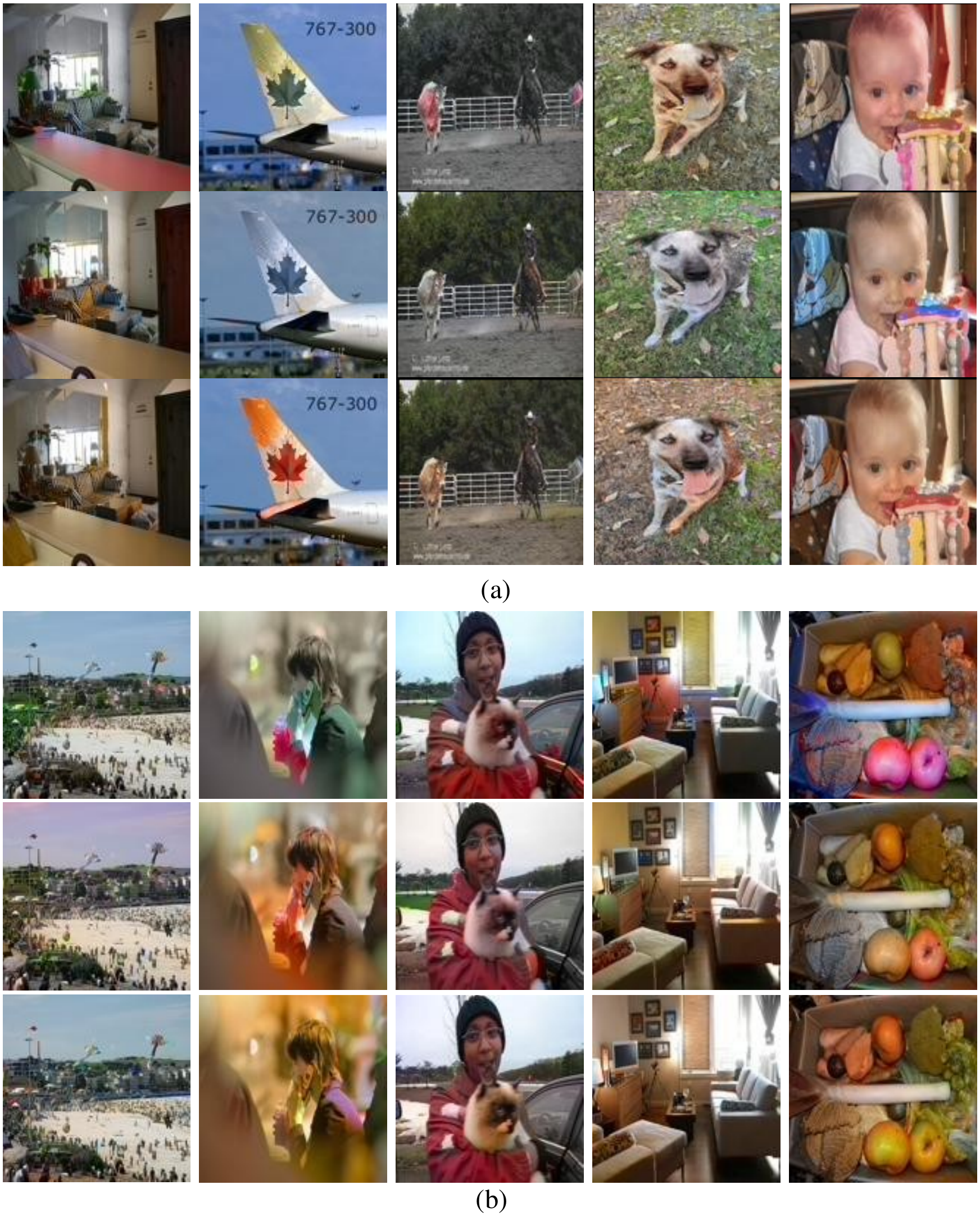}
\caption{\textbf{Colorizations produced by the generators $g^\omega$, using three variants of our network} on (a) Pascal VOC and (b) COCO-stuff: the purple flow (first row), the purple-blue flow (second row) and the purple-blue-green flow (third row). Using pixel-level semantics to guide the generator in addition to the color embedding function achieves the most realistic results.}   
\label{Fig:result2}
\end{figure*}

\subsection{Effect of segmentation on the generator $g^\omega$}
In the next experiment, we add semantics to the generator as described in Section 3.2.2 (combining the purple flow with the blue and green flows). This means the generator produces a current pixel color distribution conditioned not only on the previous colorized pixels and the color embeddings from the gray image, but also on the semantic labels. As we train the three loss functions $L^{emb}$, $L^{seg}$ and $L^{gen}$ simultaneously, we want to know whether the color embeddings produced by the embedding function are further improved. In Figure~\ref{Fig:embed2}, we compare the color embeddings generated by the embedding functions of the purple flow (shown in the top row), the purple-blue flow (shown in the second row) and the purple-blue-green flow (shown in the bottom row). Visualizations of color embeddings followed by the corresponding predicted chrominances are given. As can be seen, the addition of the green flow further improves the embedding function. From the predicted {\textbf{a}} and {\textbf{b}} visualizations, we observe better cohesion of colors for the objects. Clearly, the colorization benefits from the multi-task learning by jointly training the three different losses.        

Indeed, using semantic labels as condition to train the generator results in better color embeddings. Moreover, the final generated colorized results will also be better. In Figure~\ref{Fig:result2}, we compare the results from the three methods: pixelated colorization without semantic guidance (the purple flow), pixelated semantic color embedding (the purple-blue flow), and pixelated semantic color embedding and generator (the purple-blue-green flow). The purple flow does not always understand the object semantic well, sometimes assigning unreasonable colors to objects, such as the cow in the third example of Pascal VOC, the hands in the second example and the apples in the last example of COCO-stuff. In addition, it also suffers from inconsistency and noise on objects. Using pixelated semantics to guide the color embedding function reduces the color noise and somewhat improves the results. Adding semantic labels to guide the generator improves the results further. As shown in Figure~\ref{Fig:result2}, the purple-blue-green flow produces the most realistic and plausible results. Note that it is particularly apt at processing the details and tiny objects. For instance, the tongue of the dog is red and the lip and skin of the baby have very natural colors.   

To conclude, these experiments demonstrate our strategies using pixelated semantics for colorization are effective.

\begin{figure}
\centering
\includegraphics[scale=0.37]{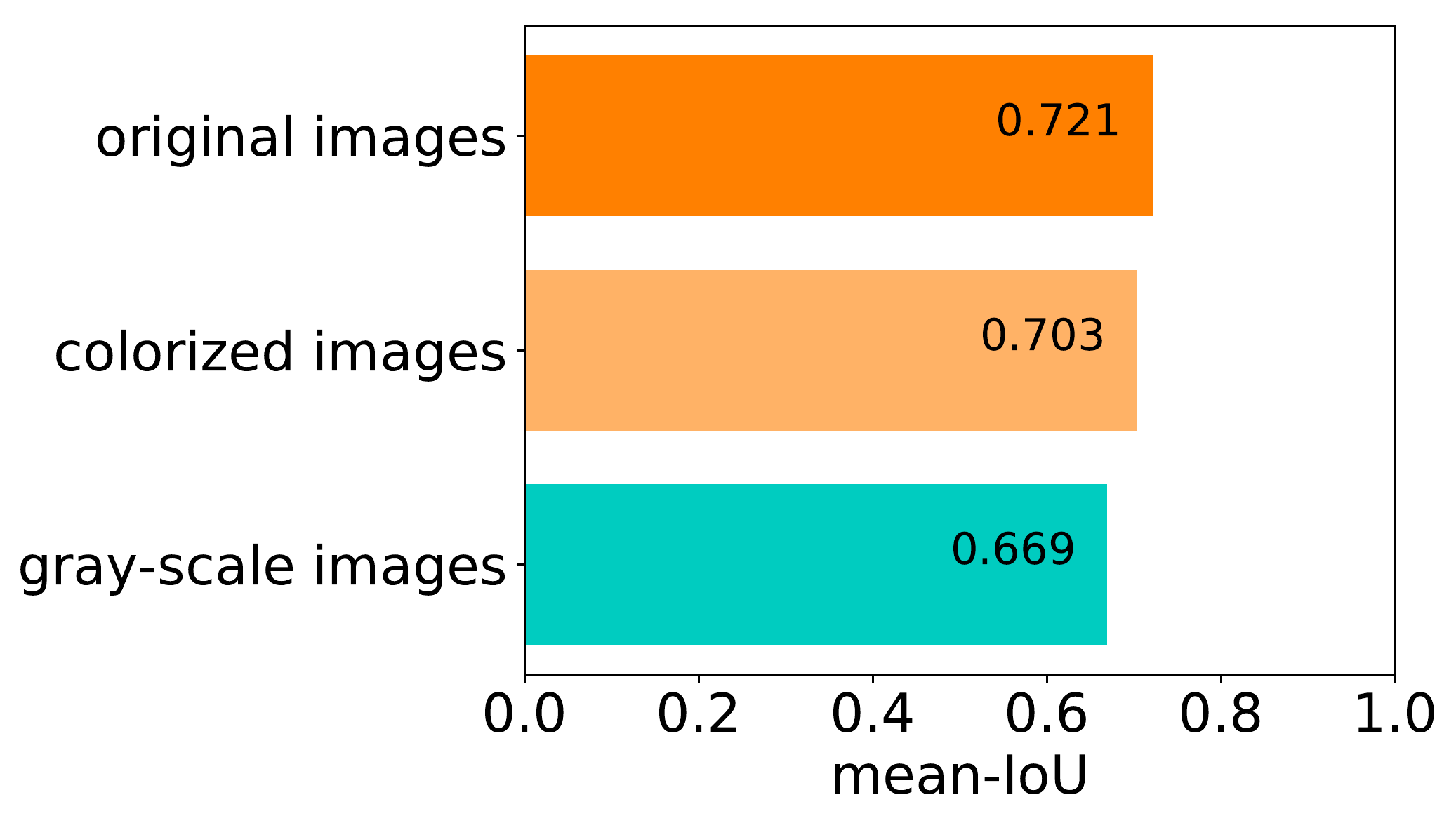}
\caption{{\textbf{Segmentation results}} in terms of mean-IoU on gray-scale images, proposed colorized images and original color images, on the Pascal VOC2012 validation dataset. Color aids semantic segmentation.}   
\label{Fig:seg1}
\end{figure}

\subsection{Effect of colorization on the segmentation} From the previous discussion, it is concluded that semantic segmentation aids in training the color embedding function and the generator. The color embedding function and the generator also help each other. As stated in Section 3, the three learnings could benefit each other. Thus, we study whether colorization is able to improve semantic segmentation. 

{\textbf{Color is important for semantic segmentation.}} As we observed in (\cite{zhao2018pixel}), color is quite critical for semantic segmentation since it captures some semantics. A simple experiment is performed to stress this point. We apply the Deeplab-ResNet101 model (\cite{chen2018deeplab}) without conditional random field as post-processing, trained on the Pascal VOC2012 training set for semantic segmentation. We test three versions of the validation images, including gray-scale images, original color images and our colorized images. The mean intersection over union (mean-IoU) is adopted to evaluate the segmentation results.  As seen in Figure~\ref{Fig:seg1}, with the original color information,  the accuracy of 72.1\% is much better than the 66.9\% accuracy of the gray images.  The accuracy obtained using our proposed colorized images is only 1.8\% lower than using the original RGB images. This again demonstrates that our colorized images are realistic. More importantly, the proposed colorized images outperform the gray-scale images by 3.4\%, which further supports the importance of color for semantic understanding.

\begin{figure}
\centering
\includegraphics[scale=0.5]{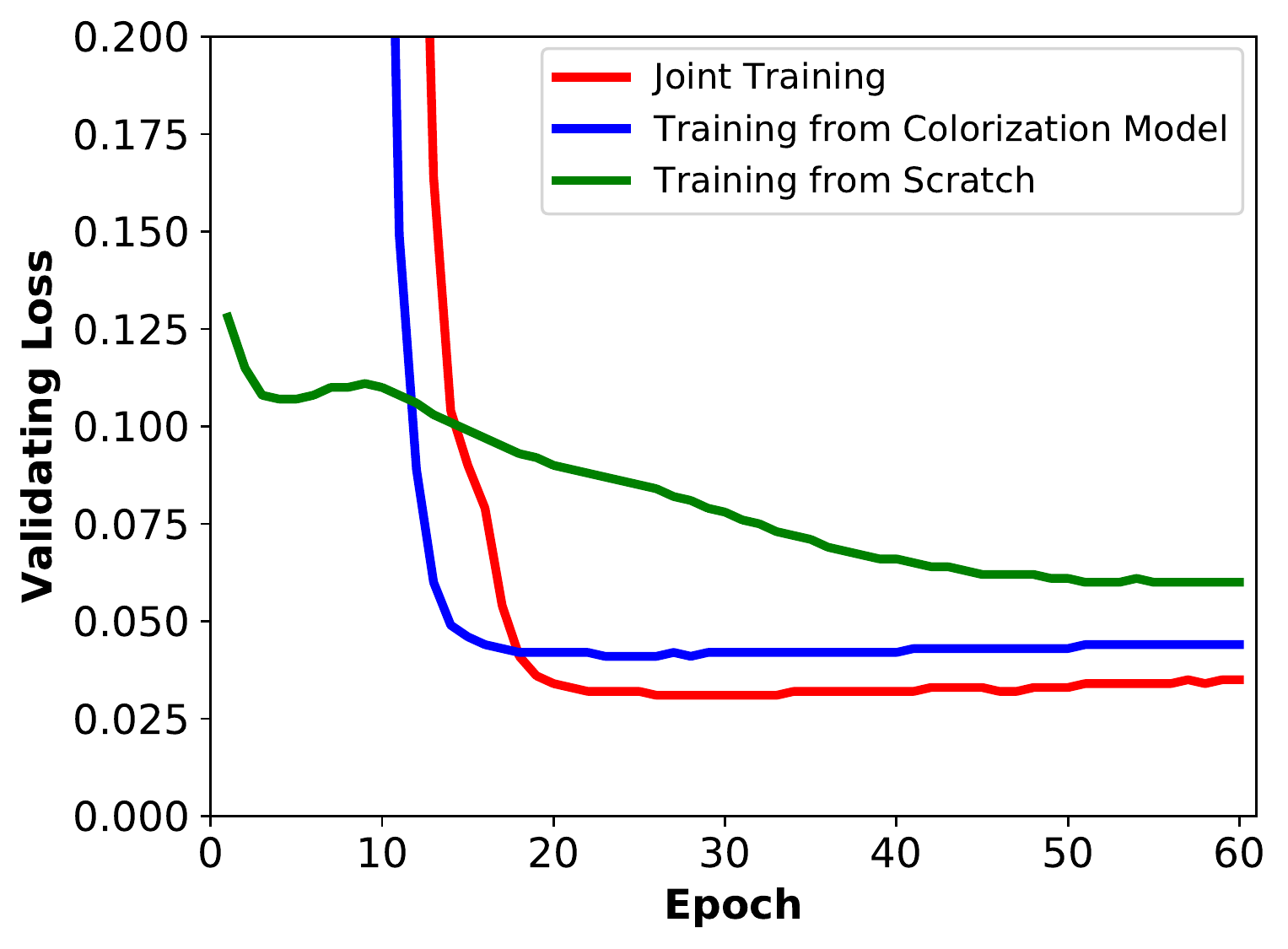}
\caption{{\textbf{Semantic segmentation validating loss comparisons}}. Three models are trained for 50 epochs. Training from a pre-trained colorization model is better than training from scratch. Jointly training obtains the lowest validating loss, which demonstrates colorization helps to improve semantic segmentation.  }   
\label{Fig:seg2}
\end{figure}

{\textbf{Colorization helps semantic segmentation.}} In order to illustrate how colorization influences semantic segmentation, we train three semantic segmentation models on gray-scale images using our network structure: (1) We jointly train semantic segmentation and colorization; (2) we only train semantic segmentation from a pre-trained colorization model; (3) we only train semantic segmentation from scratch. We train all models on the training set of Pascal VOC 2012 and test them on the validation set. As validating loss reflects the semantic segmentation accuracy on the validation set, we compare the validating loss of the three models. 

As seen in Figure~\ref{Fig:seg2}, the model trained on a pre-trained colorization model converges first. The loss is stable from the 18-th epoch and the stable loss value is about 0.043. The model trained from scratch has the lowest starting loss but converges very slowly. Starting from the 55-th epoch, the loss plateaus at 0.060. As expected, the pre-trained colorization model makes semantic segmentation achieve better accuracy. We believe the colorization model has already learned some semantic information from the colors, as also observed by \cite{zhang2016colorful}. Further, our multi-task model jointly trained with semantic segmentation and colorization obtains the lowest validating loss of 0.030, around the 25-th epoch. This supports our statement that the two tasks with the three loss functions are able to be learned harmoniously and benefit each other.      

\begin{figure}
\centering
\includegraphics[scale=0.5]{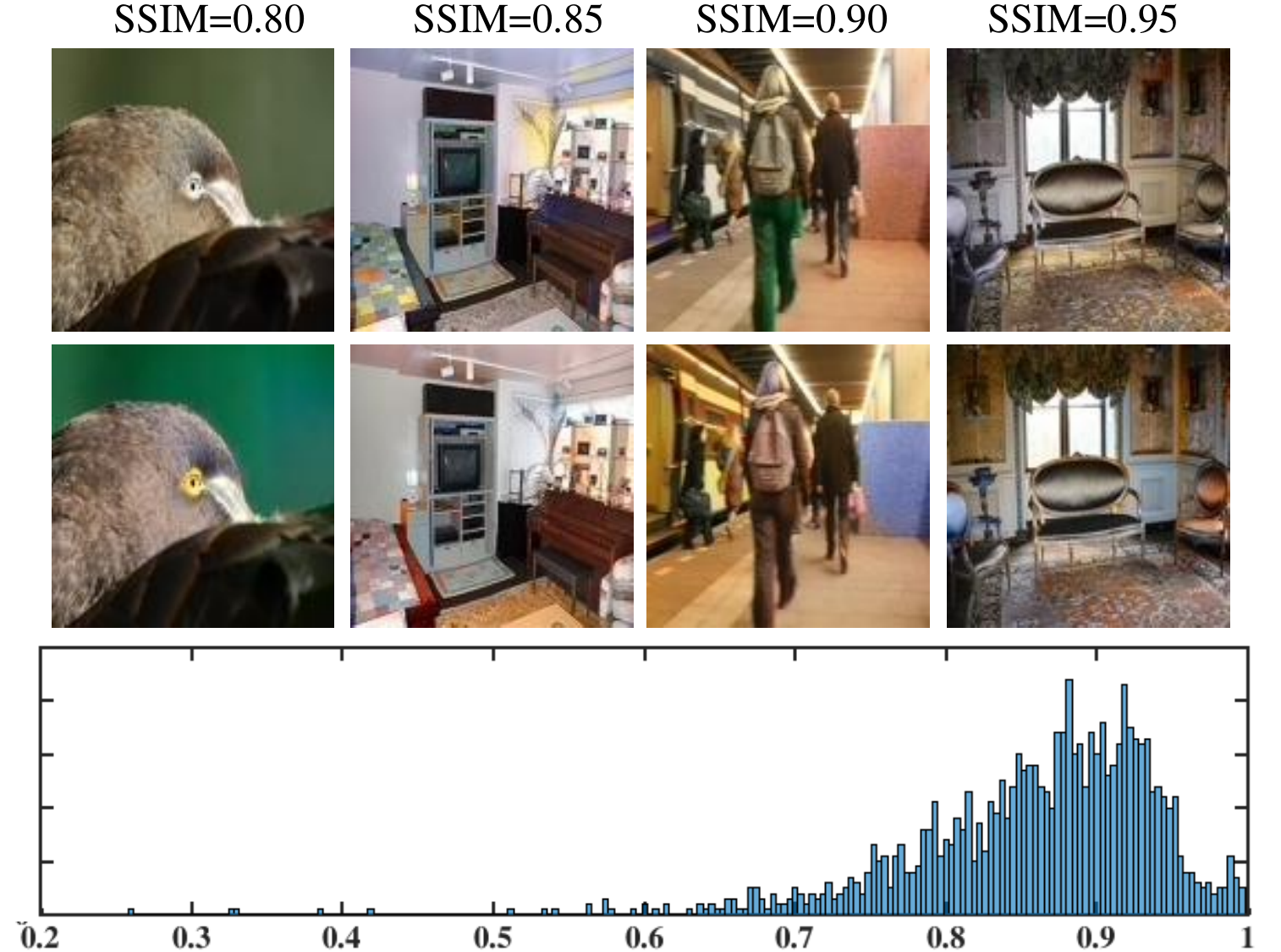}
\caption{{\textbf{Samples diversity}}. Histogram of SSIM scores on the Pascal VOC validation dataset shows the diversity of the multiple colorized results. Some examples with their specific SSIM scores are also shown. Our model is able to produce appealing and diverse colorizations.}   
\label{Fig:diversity}
\end{figure}

\begin{figure*}[t!]
\centering
\includegraphics[scale=0.78]{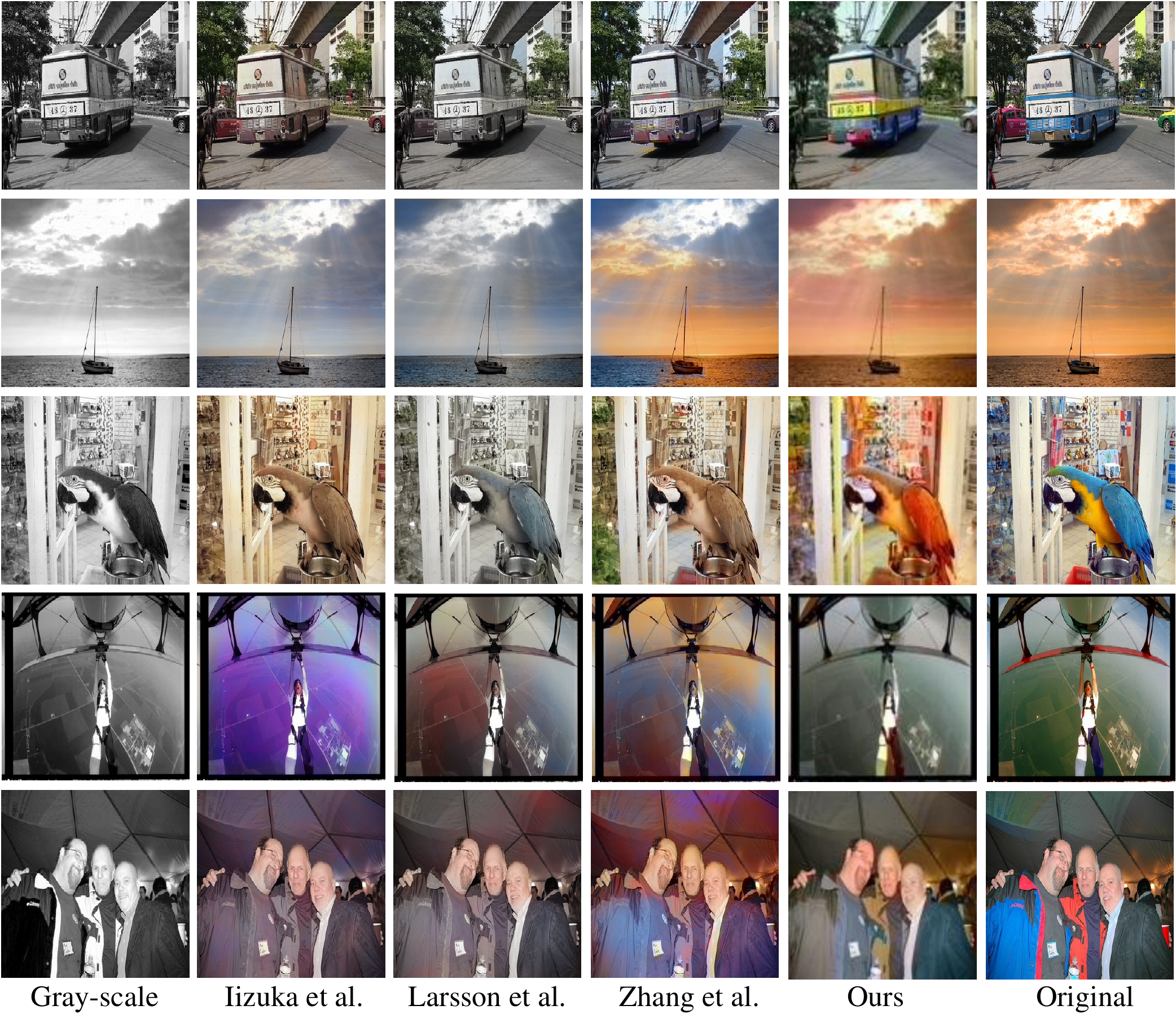}
\caption{\textbf{Comparisons with single colorization state-of-the-art}. Our results look more saturated and realistic.}   
\label{Fig:single}
\end{figure*}

\subsection{Sample Diversity}

As our model is capable of producing diverse colorization results for one gray-scale input, it is of interest to know whether or not pixelated semantics reduces the sample diversity. Following \cite{guadarrama2017pixcolor}, we compare two outputs from the same gray-scale image with multiscale structural similarity (SSIM) (\cite{wang2003multiscale}). We draw the distribution of SSIM scores for all the compared pairs on the Pascal VOC validation dataset. As shown in Figure~\ref{Fig:diversity}, most of the output pairs have an SSIM score between $0.8\sim0.95$. The examples shown in the figure demonstrate the pairs have the same content but different colors for details, such as the eyes of the bird and the pants of the lady. Usually, the large backgrounds or objects with different colors in a pair of outputs cause lower SSIM scores. For instance, the backgrounds and birds in the first example. We believe pixelated semantics does not destroy the sample diversity. We will show more diverse colorization results in the next section.


\begin{table}[t!]
	\renewcommand\arraystretch{1.1}
	\centering
	\resizebox{\columnwidth}{!}{%
		\begin{tabular}{lrr}
		\toprule
		& {\textbf{mean-IoU (\%)}} & {\textbf{PSNR(dB)}} \\ 
			\midrule

           \cite{iizuka2016let}  & 67.6 & 24.20 \\
           \cite{larsson2016learning}  & 68.8 & {\textbf{24.56}} \\
           \cite{zhang2016colorful}  & 68.1 & 22.81 \\
           Ours   & {\textbf{70.3}} & 23.15 \\
           \midrule
            \it{Ground-truth (color)} & \it{72.1} & NA \\
		\bottomrule	
		\end{tabular}
		}
	\caption{\textbf{Quantitative evaluation}. Comparisons of semantic segmentation accuracies, and PSNRs between colorized results and the ground-truth, on the Pascal VOC validation set. Our method performs better according to the mean-IoU value.}
	\label{tab:quanti}
\end{table}


\begin{table}[t!]
	\renewcommand\arraystretch{1}
	\centering
		\begin{tabular}{lr}
		\toprule
	     & {\textit{Naturalness (\%)}} \\ 

		\midrule
\rowcolor{Gray}
        \textbf{Single colorization} &\\
           \cite{iizuka2016let}  & 88.61 \\
           \cite{larsson2016learning} & 86.99 \\
           \cite{zhang2016colorful}  & 88.66  \\
           
\midrule
\rowcolor{Gray}
\textbf{Diverse colorization} &\\
			\cite{deshpande2017learning}  & 75.30 \\
			\cite{cao2017unsupervised} & 80.00\\
			\cite{ame2017bmvc} & 89.89 \\
			Ours & \bf{94.65}\\
\midrule
            \it{Ground-truth} & \it{99.58}\\
			\bottomrule
		\end{tabular}
	\caption{\textbf{Qualitative evaluation}. Comparisons of the naturalness. Our colorizations are more natural than others.}
	\label{tab:quali}
\end{table}

\begin{figure*}[t!]
\centering
\includegraphics[scale=0.65]{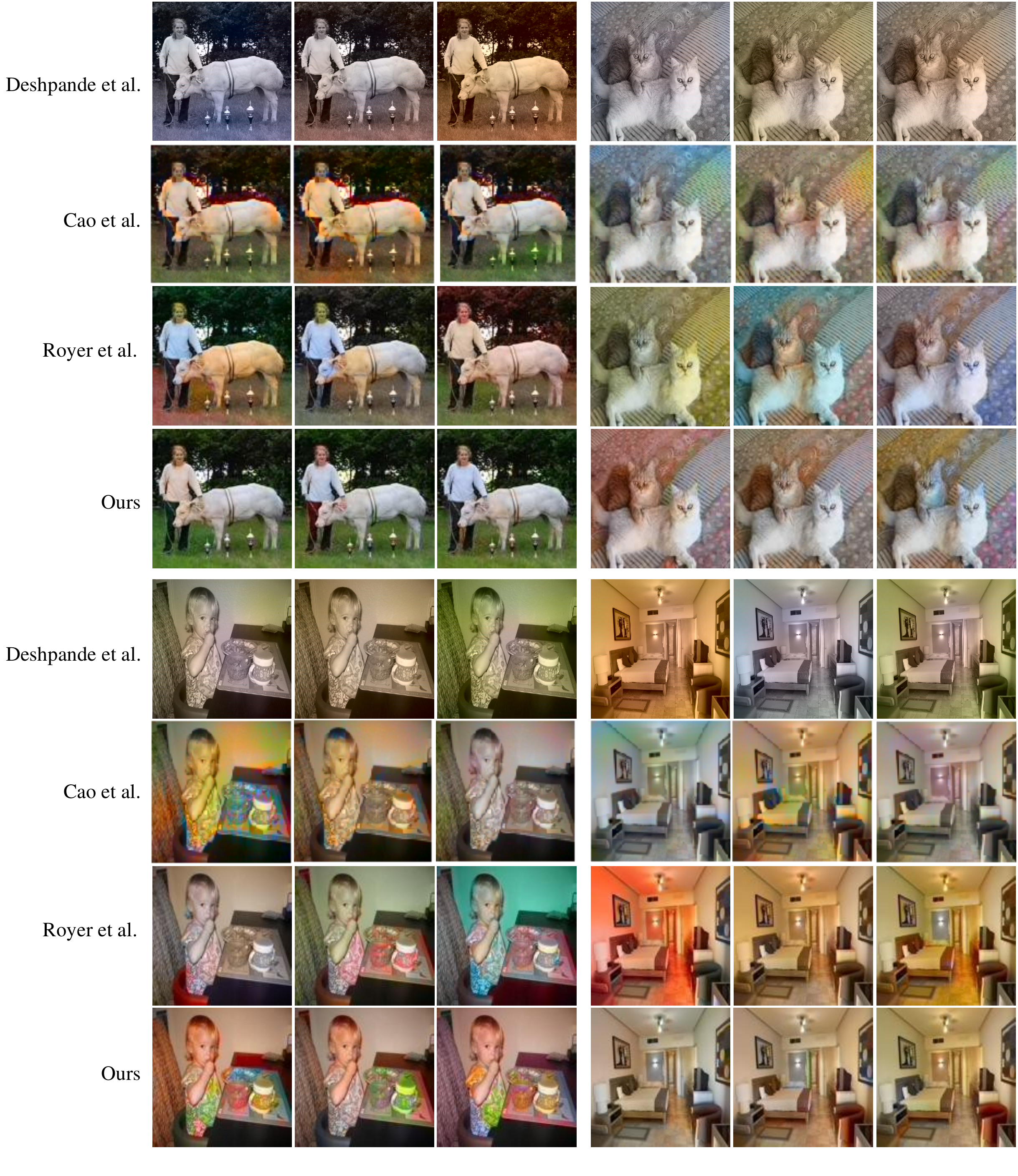}
\caption{\textbf{Comparisons with diverse colorization state-of-the-art}. The diverse results generated by our method look fairly good.}   
\label{Fig:div}
\end{figure*}
\subsection{Comparisons with State-of-the-art }

Generally, we want to produce visually compelling colorization results, which can fool a human observer, rather than recover the ground-truth. As discussed previously, colorization is a subjective challenge. Thus, both qualitative and quantitative evaluations are difficult. As for quantitative evaluation, some papers (\cite{zhang2016colorful,iizuka2016let}) apply {\textit{Top-5}} and/or {\textit{Top-1}} classification accuracies after colorization to assess the performance of the methods. Other papers (\cite{he2018deep,larsson2016learning}) use the peak signal-to-noise ratio (PSNR), although it is not a suitable criteria for colorization, especially not for a method like ours, which produces multiple results. For qualitative evaluation, human observation is mostly used (\cite{zhang2016colorful,iizuka2016let,he2018deep,ame2017bmvc,cao2017unsupervised}).  

In this paper, we propose a new evaluation method. We use semantic segmentation accuracy to assess the performance of each method, since we know semantics is key to colorization. This is more strict than classification accuracies. Specifically, we calculate the mean-IoU for semantic segmentation results from the colorized images. We use this procedure to compare our method with single colorization methods. For qualitative evaluation, we use the method from our previous work (\cite{zhao2018pixel}). We ask 20 human observers, including research students and people without any image processing knowledge, to do a test on a combined dataset including  the  Pascal VOC2012 validation and the COCO-stuff subset. Given a colorized image or the real ground-truth image, the observers should decide whether it looks natural or not.   
\subsubsection{Single Colorization State-of-the-art}

We compare the proposed method with the single colorization state-of-the-art (\cite{zhang2016colorful,iizuka2016let,larsson2016learning}). In addition to the proposed semantic segmentation accuracy evaluation, we also report PSNR. We use the Deeplab-ResNet101 model again for semantic segmentation. In this case, we only sample one result for each input, using our method.

Result comparisons are shown in Table~\ref{tab:quanti}. Our method has a lower PSNR than \cite{iizuka2016let} and \cite{larsson2016learning}, as PSNR depends on the ground-truth. PSNR over-penalizes a plausible but different colorization (\cite{he2018deep}). However, our method outperforms all the others in semantic segmentation accuracy. This demonstrates that our colorizations are more realistic and contain more perceptual semantics.

For qualitative comparison, we report the naturalness of each method according to 20 human observations in Table~\ref{tab:quali}. Three of the single colorization methods perform comparatively. Our results are more natural. We select examples are shown in Figure~\ref{Fig:single}. The method by \cite{iizuka2016let} produces good results, but sometimes assigns unsuitable colors to objects, like the earth in the fourth example. The results from \cite{larsson2016learning} look somewhat grayish. \cite{zhang2016colorful}'s method can generate saturated results but suffers from color pollution. Compared to these, our colorizations are spatially coherent and visually appealing. For instance, the color of the bird in the third example and the skin of the human in the last example, both look very natural. 

\subsubsection{Diverse Colorization State-of-the-art}

We also compare our method with the diverse colorization state-of-the art (\cite{ame2017bmvc,cao2017unsupervised,deshpande2017learning}). All of these are based on a generative model. We only qualitatively compare these by human observation. We use each model to produce three colorized samples. We report the results in Table~\ref{tab:quali}. ~\cite{ame2017bmvc} apply PixelCNN to get natural images. Our results are even more natural. Several examples are shown in Figure~\ref{Fig:div}. \cite{deshpande2017learning}, using a VAE, generate sepia toned results. \cite{cao2017unsupervised}, applying a GAN, output plausible results but with mixed colors. ~\cite{ame2017bmvc} also produces saturated results but with color pollution. Our generated colored images have fine-grained and vibrant colors and look realistic.

\subsection{Failure Cases }

\begin{figure}[t!]
\centering
\includegraphics[scale=0.6]{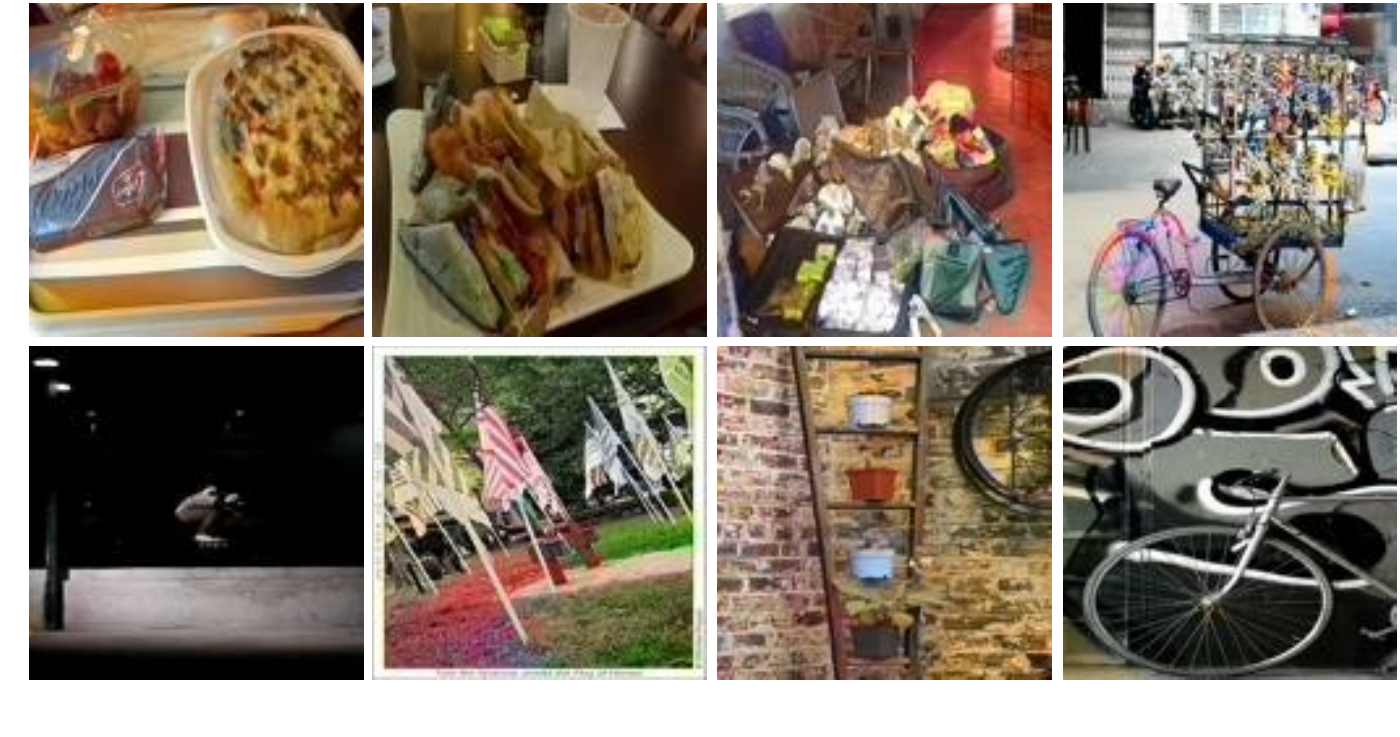}
\caption{{\textbf{Failure cases}}. Food, tiny objects and artificial objects are still very challenging.}   
\label{Fig:fail}
\end{figure}

Our method is able to output realistic colorized images but it is not perfect. There are still some
failure cases encountered by the proposed approach as well as other automatic systems. We provide a few failure cases in Figure~\ref{Fig:fail}. Usually, it is highly challenging to colorize different kinds of food. They are artificial and variable. It is also difficult to learn the semantics of images containing several tiny and occluded objects. Moreover, our method cannot handle the objects with unclear semantics. Although we exploit semantics for improving colorization, we do not have very many categories. We believe a finer semantic segmentation with more class labels will further enhance the results.

\section{Conclusion}
We propose pixelated semantic colorization to address a limitation of automatic colorization: object color inconsistency due to limited semantic understanding. We study how to effectively use pixelated semantics to achieve good colorization. Specifically, we design a pixelated semantic color embedding and a pixelated semantic generator. Both of these strengthen semantic understanding so that content confusion can be reduced. We train our network to jointly optimize colorization and semantic segmentation. The final colorized results on two datasets demonstrate the proposed strategies generate plausible, realistic and diverse colored images. Although we have achieved good results, our system is not perfect yet and has some challenges remaining. For instance, it cannot well process images with artificial objects, like food, or tiny objects. More learning examples and finer semantic segmentation may further improve the colorization results in the future.   


\bibliographystyle{spbasic}      
\bibliography{psc}   

\begin{thebibliography}{56}
\providecommand{\natexlab}[1]{#1}
\providecommand{\url}[1]{{#1}}
\providecommand{\urlprefix}{URL }
\expandafter\ifx\csname urlstyle\endcsname\relax
  \providecommand{\doi}[1]{DOI~\discretionary{}{}{}#1}\else
  \providecommand{\doi}{DOI~\discretionary{}{}{}\begingroup
  \urlstyle{rm}\Url}\fi
\providecommand{\eprint}[2][]{\url{#2}}

\bibitem[{Bugeau et~al(2014)Bugeau, Ta, and Papadakis}]{bugeau2014variational}
Bugeau A, Ta VT, Papadakis N (2014) Variational exemplar-based image
  colorization. TIP 23(1):298--307

\bibitem[{Caesar et~al(2018)Caesar, Uijlings, and Ferrari}]{caesar2018cvpr}
Caesar H, Uijlings J, Ferrari V (2018) Coco-stuff: Thing and stuff classes in
  context. In: CVPR

\bibitem[{Cao et~al(2017)Cao, Zhou, Zhang, and Yu}]{cao2017unsupervised}
Cao Y, Zhou Z, Zhang W, Yu Y (2017) Unsupervised diverse colorization via
  generative adversarial networks. In: Joint European Conference on Machine
  Learning and Knowledge Discovery in Databases

\bibitem[{Charpiat et~al(2008)Charpiat, Hofmann, and
  Sch{\"o}lkopf}]{charpiat2008automatic}
Charpiat G, Hofmann M, Sch{\"o}lkopf B (2008) Automatic image colorization via
  multimodal predictions. In: ECCV

\bibitem[{Chen et~al(2018)Chen, Papandreou, Kokkinos, Murphy, and
  Yuille}]{chen2018deeplab}
Chen LC, Papandreou G, Kokkinos I, Murphy K, Yuille AL (2018) Deeplab: Semantic
  image segmentation with deep convolutional nets, atrous convolution, and
  fully connected crfs. TPAMI 40(4):834--848

\bibitem[{Cheng et~al(2015)Cheng, Yang, and Sheng}]{cheng2015deep}
Cheng Z, Yang Q, Sheng B (2015) Deep colorization. In: ICCV

\bibitem[{Chia et~al(2011)Chia, Zhuo, Gupta, Tai, Cho, Tan, and
  Lin}]{chia2011semantic}
Chia AYS, Zhuo S, Gupta RK, Tai YW, Cho SY, Tan P, Lin S (2011) Semantic
  colorization with internet images. TOG 30(6):156

\bibitem[{Comaniciu and Meer(1997)}]{comaniciu1997robust}
Comaniciu D, Meer P (1997) Robust analysis of feature spaces: color image
  segmentation. In: CVPR

\bibitem[{Dai et~al(2016)Dai, Li, He, and Sun}]{dai2016r}
Dai J, Li Y, He K, Sun J (2016) R-fcn: Object detection via region-based fully
  convolutional networks. In: NIPS

\bibitem[{Danelljan et~al(2014)Danelljan, Shahbaz~Khan, Felsberg, and Van~de
  Weijer}]{danelljan2014adaptive}
Danelljan M, Shahbaz~Khan F, Felsberg M, Van~de Weijer J (2014) Adaptive color
  attributes for real-time visual tracking. In: CVPR

\bibitem[{Deshpande et~al(2017)Deshpande, Lu, Yeh, Chong, and
  Forsyth}]{deshpande2017learning}
Deshpande A, Lu J, Yeh MC, Chong MJ, Forsyth D (2017) Learning diverse image
  colorization. In: CVPR

\bibitem[{Everingham et~al(2015)Everingham, Eslami, Van~Gool, Williams, Winn,
  and Zisserman}]{everingham2015pascal}
Everingham M, Eslami SA, Van~Gool L, Williams CK, Winn J, Zisserman A (2015)
  The pascal visual object classes challenge: A retrospective. IJCV
  111(1):98--136

\bibitem[{Frans(2017)}]{frans2017outline}
Frans K (2017) Outline colorization through tandem adversarial networks. arXiv
  preprint arXiv:170408834

\bibitem[{Gijsenij et~al(2010)Gijsenij, Gevers, and van~de
  Weijer}]{GijsenijIJCV10}
Gijsenij A, Gevers T, van~de Weijer J (2010) Generalized gamut mapping using
  image derivative structures for color constancy. IJCV 86(2--3):127--139

\bibitem[{Guadarrama et~al(2017)Guadarrama, Dahl, Bieber, Norouzi, Shlens, and
  Murphy}]{guadarrama2017pixcolor}
Guadarrama S, Dahl R, Bieber D, Norouzi M, Shlens J, Murphy K (2017) Pixcolor:
  Pixel recursive colorization. In: BMVC

\bibitem[{Gupta et~al(2012)Gupta, Chia, Rajan, Ng, and
  Zhiyong}]{gupta2012image}
Gupta RK, Chia AYS, Rajan D, Ng ES, Zhiyong H (2012) Image colorization using
  similar images. In: Multimedia

\bibitem[{He et~al(2016)He, Zhang, Ren, and Sun}]{he2016deep}
He K, Zhang X, Ren S, Sun J (2016) Deep residual learning for image
  recognition. In: CVPR

\bibitem[{He et~al(2018)He, Chen, Liao, Sander, and Yuan}]{he2018deep}
He M, Chen D, Liao J, Sander PV, Yuan L (2018) Deep exemplar-based
  colorization. TOG 37(4):47

\bibitem[{Van~der Horst and Bouman(1969)}]{van1969spatiotemporal}
Van~der Horst GJ, Bouman MA (1969) Spatiotemporal chromaticity discrimination.
  JOSA 59(11):1482--1488

\bibitem[{Huang et~al(2005)Huang, Tung, Chen, Wang, and Wu}]{huang2005adaptive}
Huang YC, Tung YS, Chen JC, Wang SW, Wu JL (2005) An adaptive edge detection
  based colorization algorithm and its applications. In: Multimedia

\bibitem[{Iizuka et~al(2016)Iizuka, Simo-Serra, and Ishikawa}]{iizuka2016let}
Iizuka S, Simo-Serra E, Ishikawa H (2016) Let there be color!: joint end-to-end
  learning of global and local image priors for automatic image colorization
  with simultaneous classification. TOG 35(4):110

\bibitem[{Ironi et~al(2005)Ironi, Cohen-Or, and
  Lischinski}]{ironi2005colorization}
Ironi R, Cohen-Or D, Lischinski D (2005) Colorization by example. In: Rendering
  Techniques

\bibitem[{Isola et~al(2017)Isola, Zhu, Zhou, and Efros}]{isola2017image}
Isola P, Zhu JY, Zhou T, Efros AA (2017) Image-to-image translation with
  conditional adversarial networks. In: CVPR

\bibitem[{Khan et~al(2009)Khan, Van De~Weijer, and Vanrell}]{khan2009top}
Khan FS, Van De~Weijer J, Vanrell M (2009) Top-down color attention for object
  recognition. In: ICCV

\bibitem[{Khan et~al(2012)Khan, Anwer, Van~de Weijer, Bagdanov, Vanrell, and
  Lopez}]{khan2012color}
Khan FS, Anwer RM, Van~de Weijer J, Bagdanov AD, Vanrell M, Lopez AM (2012)
  Color attributes for object detection. In: CVPR

\bibitem[{Kingma and Ba(2015)}]{kingma2014adam}
Kingma DP, Ba J (2015) Adam: A method for stochastic optimization. In: ICLR

\bibitem[{Kingma and Welling(2014)}]{kingma2013auto}
Kingma DP, Welling M (2014) Auto-encoding variational bayes. ICLR

\bibitem[{Larsson et~al(2016)Larsson, Maire, and
  Shakhnarovich}]{larsson2016learning}
Larsson G, Maire M, Shakhnarovich G (2016) Learning representations for
  automatic colorization. In: ECCV

\bibitem[{Levin et~al(2004)Levin, Lischinski, and
  Weiss}]{levin2004colorization}
Levin A, Lischinski D, Weiss Y (2004) Colorization using optimization. TOG
  23(3):689--694

\bibitem[{Lin et~al(2014)Lin, Maire, Belongie, Hays, Perona, Ramanan,
  Doll{\'a}r, and Zitnick}]{lin2014microsoft}
Lin TY, Maire M, Belongie S, Hays J, Perona P, Ramanan D, Doll{\'a}r P, Zitnick
  CL (2014) Microsoft coco: Common objects in context. In: ECCV

\bibitem[{Liu et~al(2008)Liu, Wan, Qu, Wong, Lin, Leung, and
  Heng}]{liu2008intrinsic}
Liu X, Wan L, Qu Y, Wong TT, Lin S, Leung CS, Heng PA (2008) Intrinsic
  colorization. TOG 27(5):152

\bibitem[{Long et~al(2015)Long, Shelhamer, and Darrell}]{long2015fully}
Long J, Shelhamer E, Darrell T (2015) Fully convolutional networks for semantic
  segmentation. In: CVPR

\bibitem[{Lou et~al(2015)Lou, Gevers, Hu, Lucassen et~al}]{lou2015color}
Lou Z, Gevers T, Hu N, Lucassen MP, et~al (2015) Color constancy by deep
  learning. In: BMVC

\bibitem[{Luan et~al(2007)Luan, Wen, Cohen-Or, Liang, Xu, and
  Shum}]{luan2007natural}
Luan Q, Wen F, Cohen-Or D, Liang L, Xu YQ, Shum HY (2007) Natural image
  colorization. In: Proceedings of the 18th Eurographics conference on
  Rendering Techniques

\bibitem[{Noh et~al(2015)Noh, Hong, and Han}]{noh2015learning}
Noh H, Hong S, Han B (2015) Learning deconvolution network for semantic
  segmentation. In: ICCV

\bibitem[{van~den Oord et~al(2016)van~den Oord, Kalchbrenner, Espeholt,
  Vinyals, Graves et~al}]{van2016conditional}
van~den Oord A, Kalchbrenner N, Espeholt L, Vinyals O, Graves A, et~al (2016)
  Conditional image generation with pixelcnn decoders. In: NIPS

\bibitem[{Perez et~al(2018)Perez, Strub, De~Vries, Dumoulin, and
  Courville}]{perez2017film}
Perez E, Strub F, De~Vries H, Dumoulin V, Courville A (2018) Film: Visual
  reasoning with a general conditioning layer. In: AAAI

\bibitem[{P{\'e}rez et~al(2002)P{\'e}rez, Hue, Vermaak, and
  Gangnet}]{perez2002color}
P{\'e}rez P, Hue C, Vermaak J, Gangnet M (2002) Color-based probabilistic
  tracking. In: ECCV

\bibitem[{Polyak and Juditsky(1992)}]{polyak1992acceleration}
Polyak BT, Juditsky AB (1992) Acceleration of stochastic approximation by
  averaging. SIAM Journal on Control and Optimization 30(4):838--855

\bibitem[{Qu et~al(2006)Qu, Wong, and Heng}]{qu2006manga}
Qu Y, Wong TT, Heng PA (2006) Manga colorization. TOG 25(3):1214--1220

\bibitem[{Radford et~al(2016)Radford, Metz, and
  Chintala}]{radford2015unsupervised}
Radford A, Metz L, Chintala S (2016) Unsupervised representation learning with
  deep convolutional generative adversarial networks. In: ICLR

\bibitem[{Royer et~al(2017)Royer, Kolesnikov, and Lampert}]{ame2017bmvc}
Royer A, Kolesnikov A, Lampert CH (2017) Probabilistic image colorization. In:
  BMVC

\bibitem[{Salimans et~al(2017)Salimans, Karpathy, Chen, Kingma, and
  Bulatov}]{salimans2017pixelcnn++}
Salimans T, Karpathy A, Chen X, Kingma DP, Bulatov Y (2017) Pixelcnn++: A
  pixelcnn implementation with discretized logistic mixture likelihood and
  other modifications. In: ICLR

\bibitem[{Sanchez and Binefa(2000)}]{sanchez2000improving}
Sanchez JM, Binefa X (2000) Improving visual recognition using color
  normalization in digital video applications. In: ICME

\bibitem[{van~de Sande et~al(2010)van~de Sande, Gevers, and
  Snoek}]{SandePAMI10}
van~de Sande KEA, Gevers T, Snoek CGM (2010) Evaluating color descriptors for
  object and scene recognition. TPAMI 32(9):1582--1596

\bibitem[{Sangkloy et~al(2017)Sangkloy, Lu, Fang, Yu, and
  Hays}]{sangkloy2017scribbler}
Sangkloy P, Lu J, Fang C, Yu F, Hays J (2017) Scribbler: Controlling deep image
  synthesis with sketch and color. In: CVPR

\bibitem[{Swain and Ballard(1991)}]{swain91}
Swain MJ, Ballard DH (1991) Color indexing. IJCV 7(1):11--32

\bibitem[{Tai et~al(2005)Tai, Jia, and Tang}]{tai2005local}
Tai YW, Jia JY, Tang CK (2005) Local color transfer via probabilistic
  segmentation by expectation-maximization. In: CVPR

\bibitem[{Vondrick et~al(2018)Vondrick, Shrivastava, Fathi, Guadarrama, and
  Murphy}]{vondrick2018tracking}
Vondrick C, Shrivastava A, Fathi A, Guadarrama S, Murphy K (2018) Tracking
  emerges by colorizing videos. In: ECCV

\bibitem[{Wang et~al(2018)Wang, Yu, Dong, and Loy}]{wang2018recovering}
Wang X, Yu K, Dong C, Loy CC (2018) Recovering realistic texture in image
  super-resolution by deep spatial feature transform. In: CVPR

\bibitem[{Wang et~al(2003)Wang, Simoncelli, and Bovik}]{wang2003multiscale}
Wang Z, Simoncelli EP, Bovik AC (2003) Multiscale structural similarity for
  image quality assessment. In: The Thrity-Seventh Asilomar Conference on
  Signals, Systems \& Computers

\bibitem[{Welsh et~al(2002)Welsh, Ashikhmin, and
  Mueller}]{welsh2002transferring}
Welsh T, Ashikhmin M, Mueller K (2002) Transferring color to greyscale images.
  TOG 21(3):277--280

\bibitem[{Yatziv and Sapiro(2006)}]{yatziv2006fast}
Yatziv L, Sapiro G (2006) Fast image and video colorization using chrominance
  blending. TIP 15(5):1120--1129

\bibitem[{Zhang et~al(2016)Zhang, Isola, and Efros}]{zhang2016colorful}
Zhang R, Isola P, Efros AA (2016) Colorful image colorization. In: ECCV

\bibitem[{Zhang et~al(2017)Zhang, Zhu, Isola, Geng, Lin, Yu, and
  Efros}]{zhang2017real}
Zhang R, Zhu JY, Isola P, Geng X, Lin AS, Yu T, Efros AA (2017) Real-time
  user-guided image colorization with learned deep priors. In: SIGGRAPH

\bibitem[{Zhao et~al(2018)Zhao, Liu, Snoek, Han, and Shao}]{zhao2018pixel}
Zhao J, Liu L, Snoek CGM, Han J, Shao L (2018) Pixel-level semantics guided
  image colorization. In: BMVC

\end{thebibliography}

\end{sloppypar}
\end{document}